\documentclass{article}
\usepackage{graphicx} % Required for inserting images
\title{Deep Learning-Based Robust Optical Guidance for Hypersonic Platforms}
\author{Adrien Chan-Hon-Tong, Aurélien Plyer \\ Baptiste Cadalen, Laurent Serre \\ ONERA Université Paris-Saclay \\ firstname.lastname@onera.fr}
\begin{document}
\maketitle

\begin{center}
    \textbf{Abstract}

    Sensor-based guidance is required for long-range platforms. To bypass the structural limitation of classical registration on reference image framework, we offer in this paper to encode a stack of images of the scene into a deep network. Relying on a stack is showed to be relevant on bimodal scene (e.g. when the scene can or can not be snowy).
\end{center}

\section{Introduction}
Localizing a camera using the current image is as old as computer vision \cite{smith1986representation}. However, SLAM frameworks (Simultaneous Localization and Mapping) only offer relative localization.
To restore absolute localization, one must combine the information provided by the current image with external information such as GNSS, or, anchor points (points for which the absolute 3D position is known) visible in the image.
This last idea leads to the framework of registration on a reference image widely used in remote sensing: by using anchors recognized in the current image, PnP algorithms \cite{longuet1981computer} allow restoring the absolute position of the camera (and even it related coordinate system).

However, this approach suffers from a structural drawback: it is heavily dependent on the quality of the reference image and on the similarity between this reference image and the current one.
To bypass this limitation, this paper proposes relying instead on a stack of images of the scene to capture common changes that can arise (e.g., snowy or not) and to implicitly fill in missing information in each individual image (e.g., each image may contain clouds, but the entire scene can be seen across the stack).

As manipulating the stack is inconvenient, we propose using a deep network to directly learn a mapping between the current image and the absolute position.
This approach is particularly relevant for optical guidance of hypersonic platforms.
In such contexts, GNSS can be denied, and embedded accelerometers lead to large drifts after several thousand kilometers of travel.
Then, using a stack to achieve some invariance can be straightforwardly extended to invariance to the precise spectral band and/or potential distortion created by heat when capturing the scene at very high speeds.
Finally, in this context, the absolute position is somewhat less important than the relative position with the known 3D position of the target.
This allows specializing the deep network to map the current image with the direction of the target, decreasing the number of required layers.

Despite this approach clearly introducing logistical issues (the need to collect a stack of images of the scene, the need to train a specific network for a single target, the lack of well-understood geometric foundations), we provide a case of a bimodal scene (with and without snow) where classical baseline  fails while our method mitigates the bimodal issue.

\section{Related Works}
There is a very large literature on SLAM and registration, currently being revisited by the rise of efficient deep network methods for geometric tasks.
SIFT+lightglu \cite{lindenberger2023lightglue}, which combines original SIFT \cite{lowe2004distinctive} and efficient deep learning descriptors (an idea introduced in \cite{yi2016lift}), seems to be the current state of the art of image matching, challenged by new approaches performing end-to-end dense matching \cite{sun2021loftr,edstedt2024roma,he2025matchanything}.
However, SIFT+lightglu focuses on robustness to point of view. So it may be sensitive to strong changes in the appearance of the scene. End-to-end dense matching methods may be more robust to those changes by implicitly learning the existence of such drift (MatchAnything \cite{he2025matchanything} can even match an image to a symbolic map, for example), but they are today implemented with very expensive transformer layers making them unacceptable for embedded platforms (the web demo of \cite{he2025matchanything} requires 16s per pair of small images).

Also, from a functional point of view, SIFT+lightglu and MatchAnything perform registration, not directly the final guidance task.
In this sense, appearance-only SLAM like Fab-Map\cite{cummins2008fab} is somehow related to this work.
Yet, Fab-Map aims to detect already known areas (loop closing) while we map image appearance to camera/target localization.

Let us point out that our idea of creating an implicit model of a scene from a stack of images is also related to Nerf (Neural Radiance Field) literature \cite{mildenhall2021nerf}.
However, here we do not really model the scene but rather the appearance of the target and/or it surrounding at different scales/orientations...

To summarize, our work is inspired by Nerf but applied to guidance. It does not rely on transformer-based dense correlation to perform registration, thus being much faster than MatchAnything. Finally, compared to classical registration techniques whose current state of the art seems to be SIFT+lightglu, our pipeline does not depend on a specific reference image, offering robustness to common changes in the scene.

\section{Direct Guidance Learning}
Before describing our method, let us point out that from a theoretical point of view, registration can be cast as an inverse problem. In many problems, given latent parameters $\theta$ (here the camera pose), one may generate $x = G(\theta)$ (here the image) without having the ability to do the opposite, i.e., recovering $\theta$ from $x$. One solution can be to sample a large set of parameters $\theta_1, ..., \theta_N$ and approximate $\widehat{G^{-1}}(x_n)=\theta_n$ for example with deep networks (implicitly performing some kind of interpolation between known $\theta$).
In the case of guidance, the process generates both $x$ and $y$ (here the relative target position) given the latent parameter $\theta$, and we do not need to recover the latent parameter $\theta$ but directly learn the mapping between $x$ and $y$ simplifying further the problem.
This idea is summarized in figure 1.
Given a stack of geocalized images $X_1,...,X_K$ and known target positions $p_1,...,p_k$ in each image, we sample those images $k_1,...,k_R$ and homothetic matrices $A_1,..., A_R$  representing the link between pixel $i,j$ in an image and pixel $u,v$ in reference image $k$.
Each of these matrices $A_r$ allows\footnote{Importantly, the projection related to matrix $A$ ignores 3D effects (topography would be necessary in addition to geocalized images to fully capture 3D effects). However, those effects may not be critical in remote sensing and around nadir orientation of the camera.} to create image $x_r$ corresponding to how the scene would have been viewed (at time when image $X_{k_r}$ was taken) from camera $A_r$ (rather than the camera related to $X_{k_r}$), but also $y_r$, the corresponding position of $p_{k_r}$.
\begin{figure}
\centering
\includegraphics[width=0.8\linewidth]{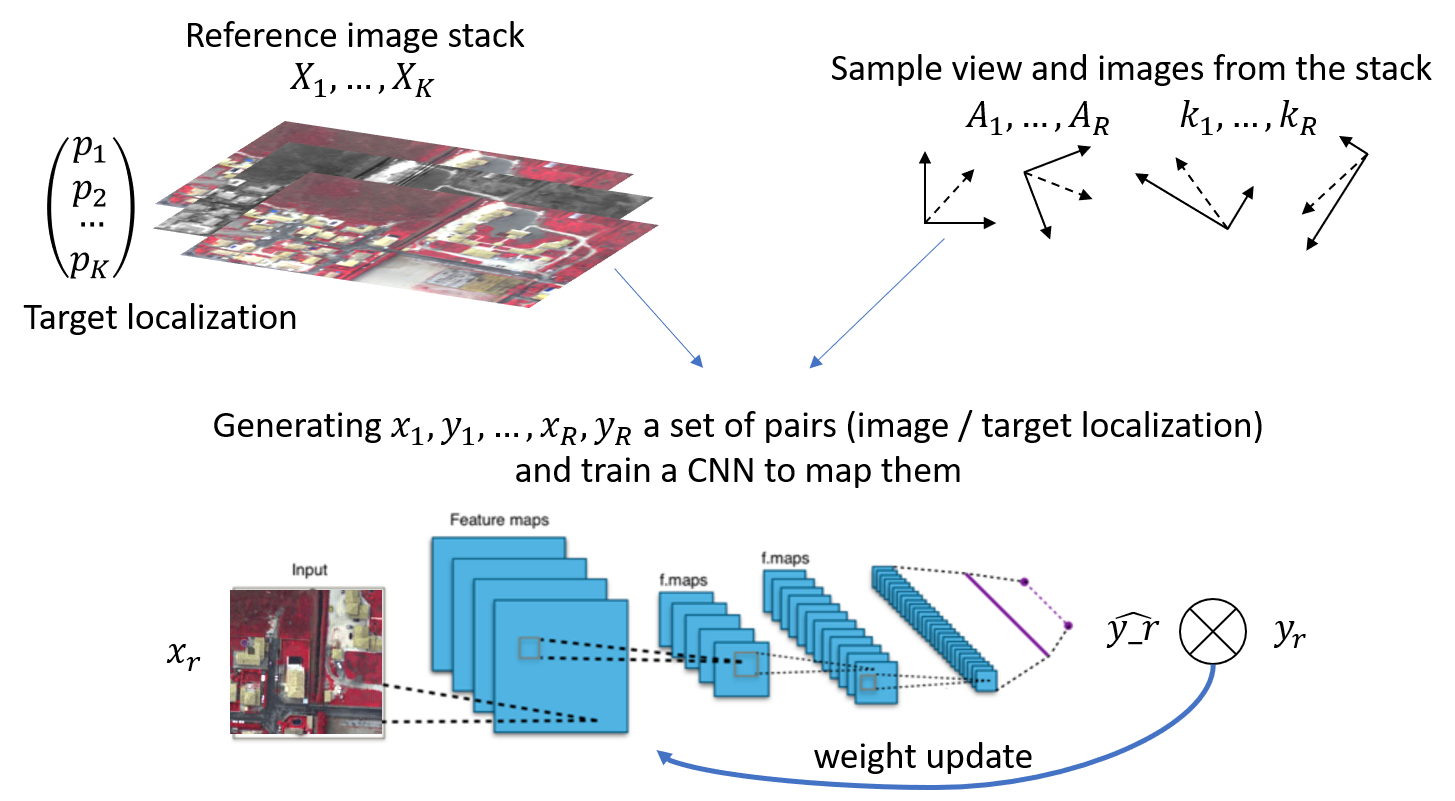}
\caption{Overview of the offered framework: guidance is cast as the problem of learning the target localization $y_r$ from an image $x_r$ - at training time, the deep network is trained by sampling views from the stack of reference images - at runtime, the network directly predicts a localization from the current image.}
\label{fig:overview}
\end{figure}
This way, we are able to create a large dataset of pairs of image/target localization $(x_1,l_1), ..., (x_R,l_R)$ from which we train a deep network $f$ with weights $w$ to predict $l_r$ from $x_r$ using simple regression loss $\mathcal{L}(w,x_r,l_r) = (f_w(x_r)-l_r)^2$ or by selecting the pixel containing the target (both ways have proven similar in our experiment with regression being more straightforward).

During inference, given the current image $x$, $f_w(x)$ directly predicts the relative localization of the target.
To assess their quality, those predictions have been evaluated on a test dataset (a set of pairs of image/target localization generated with the same process as the training set but disjoint from the training one).

\section{Experiments}
We consider 3 use cases.
First, we aim to measure performances under large diversity of camera positions. For these experiments, we sample a random rotation (around the vertical), a random zoom, and a small uniform rotation around the other two axes to generate each matrix $A_r$.
This setting is split into 2 experiments. One setting is with weak change as we consider a single large image from IGN BD Ortho to create views.
Then, we consider a setting with strong change using 4 Sentinel2 images (from January 2025 to March 2025) where 2 images contain snow and 2 do not.
The first two (1 snow, 1 no-snow) are used for training, and the other two for testing.
Thus, in this test, the deep network does neither know the point of view nor the image (and so the fact that there will or will not be snow) and implicitly needs to use the correct reference images when associating current views with the internal model encoded in the network weights.
Finally, we also consider a more representative setting where the points of view are not sampled randomly but along  trajectories of an hypersonic platform\footnote{These are not truly representative trajectories but we consider features like rotation around the main axis which are common with true ones.} and with a S2 infrared image with low change.

\subsection{Implementation Details}
As the higher we are, the harder it is to be precise on target coordinates in the metric system, we normalize using pixel coordinates: the task becomes selecting the pixel in which the target belongs (even if not visible when the platform is too high).
We thus report both mean square error and number of samples for which the error is less than 10px to avoid the metric being biased by outliers.

S2 images are clipped min-max at 0.2 percentile and gamma corrected at 0.5. In these experiments, S2 images are mostly cloudless (but with large changes in snow), yet the algorithm is designed to resist low cloud cover (obviously, images with strong cloud cover should not be added to the stack of reference images).

For the deep network, we put emphasis on limiting the number of layers: all experiments are done with the first 4 blocks of ConvNext Tiny \cite{liu2022convnet} followed by a task-specific dense layer allowing us to achieve 60FPS on CPU only at inference on 256x256 images. 
Currently, the same model with the first 4 blocks of EfficientNet B0 \cite{tan2019efficientnet} has been tested but performs poorly in comparison to ConvNext Tiny despite both networks representing the state of the art of small convolutional deep networks. Probably, EfficientNet would have required more blocks to capture the problem, damaging running time.
Let us point out that with this setting, the accuracy of the offered method cannot be higher than 5 pixels (it predicts an 8x8 pixel block containing the target).

Fine-tuning of the network pretrained on Imagenet is done in several steps: first, the head is aligned on the task; then, a first fine-tuning is performed with SGD and very small lr; finally, a classical fine-tuning is performed with advanced optimizer \cite{defazio2024road}. Let stress that pretraining weights are critically required (pretraining in S2 data would be probably allow much better results).

The baseline is opencv2 standard SIFT registration on a single reference image with standard Lowe's ratio, approximate position of the camera is provided to the baseline to help the registration (not needed with the offered method which directly map image to position). Currently we also tested SIFT+lightGlue (pretrained) but it performs similarly as SIFT: this can be explained because first lightGlue is not trained for remote sensing image, and then, because the algorithm should not try produce a very precise wrapping but rather to deal with very large appearance change, and, for this purpose, pretrained lightGlue descriptors were not more usefull than SIFT ones.

\subsection{Guidance under Weak Change}
The first setting is mostly an experiment designed to ensure algorithms are functional (see illustration figure 2):
both methods successfully manage to find efficiently the position of the target as reported in table 1.
Precisely, the SIFT baseline achieves better precision than the offered method (in this weak-change setting) in terms of mean square error, but both methods produce acceptable predictions in 96\% of the sampled images (most failures are related to images with strong oblique views which are somehow distorted by the absence of topographic data).
\begin{table}[h]
\centering
\begin{tabular}{c|c|c}
& mse & frames with error less than 10px \\\hline
SIFT baseline & \textbf{1.68px} & \textbf{96.4\%} \\\hline
Offered method & 6.58px & 96.1\%
\end{tabular}
\caption{Performance of target position estimation under weak change between reference image and current image}
\label{tab:res1}
\end{table}
\begin{figure}
    \centering
    \includegraphics[width=0.9\linewidth]{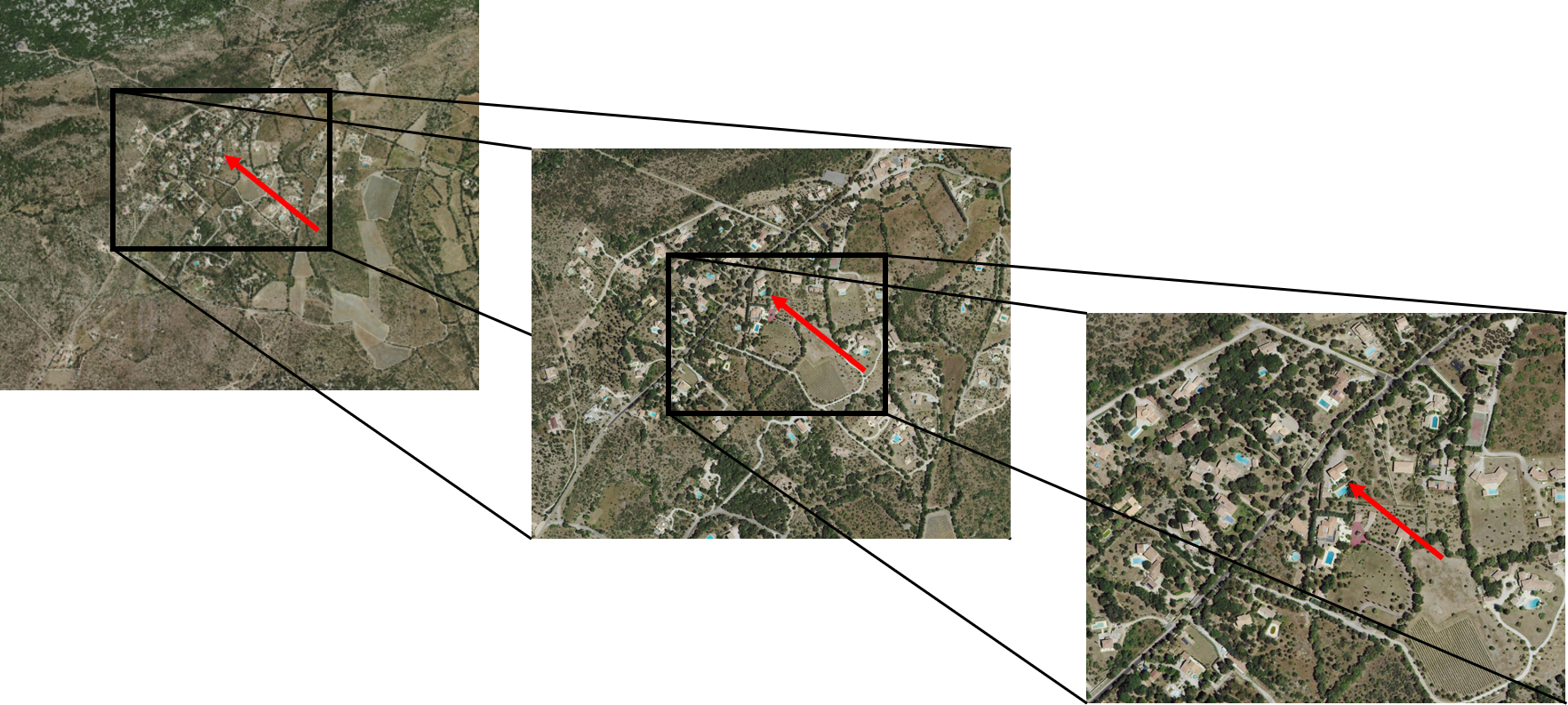}
    \caption{Illustration of the image of this first experiment. Hypothetical target (red arrow) is not really visible in first image, yet, the surrounding is sufficient to know where it is. This explains how our model is able to learn a mapping image-to-target at any resolution.}
    %\label{fig:enter-label}
\end{figure}

\subsection{Guidance under Strong Change}
For the second setting, 2 images (1 with snow, 1 without snow) are used for training the offered algorithm, and the same for testing.
However, the baseline is restricted to selecting a single reference image, making it hard to register on the opposite test image.
This leads to less than 24\% of the test images being correctly processed (in many cases, SIFT matching does not even find 4 good matches for estimating the homography matrix).
Inversely, the offered method manages to process correctly more than half of the images distributed across the two modes (snowy and non-snowy).
Currently, performance of our method on training images is much higher, highlighting the fact that performance may increase significantly with a larger reference image stack (only two here).
\begin{table}[h]
\centering
\begin{tabular}{c|c|c}
& mse & frames with error less than 10px \\\hline
SIFT baseline & 53.03px & 23.6\% \\\hline
Offered method & \textbf{42.64px} & \textbf{51.3\%}
\end{tabular}
\caption{Performance of target position estimation under strongly bimodal (snow vs no-snow) distribution of reference and testing images}
\label{tab:res2}
\end{table}

These results, reported in table 2, highlight the fact that relying on a single reference image is not a good idea when strong changes can arise between the reference image and the current one, while encoding the scene with our method on a stack of reference images can mitigate the issue.

In order to make more visual why SIFT performs poorly, figure 3 displays the same crop of two S2 one snowy, one normal (centered on an hypothetical target). One can see how the appearance are different even without any geometrical changes.
On this already-registered pair, SIFT extracts around 700 points per image but manage to match only 20 of them.
Adding only a little geometric deformation or sub-sampling frequently makes the number of matches going under 4. 
\begin{figure}
    \centering
    \includegraphics[width=0.9\linewidth]{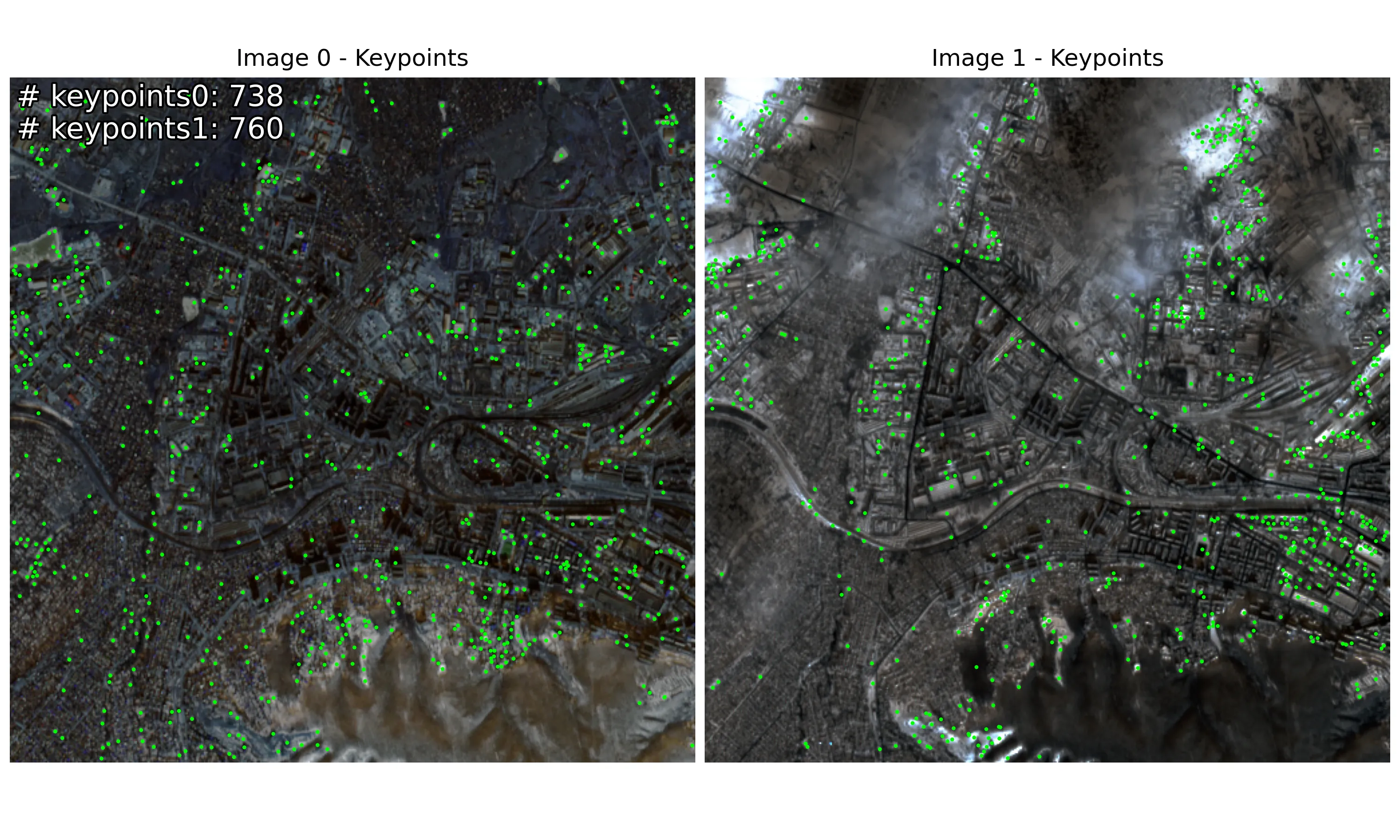}
    
    \caption{Illustration of limits of registration on a single reference image in presence of strong change: the image displays the same region in two S2 images with green dot being the SIFT keypoints. Due to important appearance change only 20 SIFT will then be matched while the pair is already registered. Adding sub-sampling or geometric deformation frequently makes SIFT unable to perform registration while the offered baseline just learn to predict the target. Illustration done with github.com/Vincentqyw/image-matching-webui.}
    %\label{fig:enter-label}
\end{figure}

\subsection{Guidance across Trajectories}
As the images seen along a trajectory way exhibit some specificity, we also offer to evaluate performance not on individual images but on videos related to trajectories of the platform.
Thus, instead of sampling views $A_r$ randomly, we simulate trajectories of a camera in the head of an hypersonic platform performing somewhat representative moves.
We simulate 100 trajectories (around the same scene/target under weak change setting like in 4.2 with an infrared S2 image), 90 for training and 10 for testing.
All images from all training trajectories are used for training the network, like for other experiments: views are considered independent, but the fact they belong to a trajectory correspond to a different sampling.

Instead of evaluating each image independently, we consider guidance successful along the trajectory if the predicted target is correct within 10px on at least 66\% of the frames (without 4-consecutive wrong frames).
With this definition, we manage to have successful guidance on all 10 testing trajectories.

Figures 4 displays output of the algorithm along a testing trajectory.
\begin{figure}
    \centering
    \includegraphics[width=0.23\linewidth]{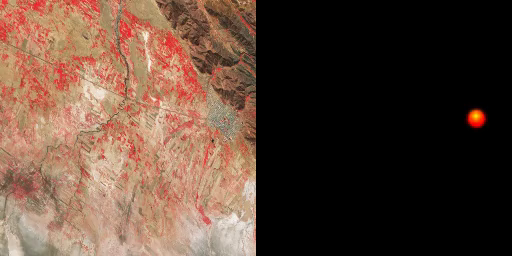}
    \includegraphics[width=0.23\linewidth]{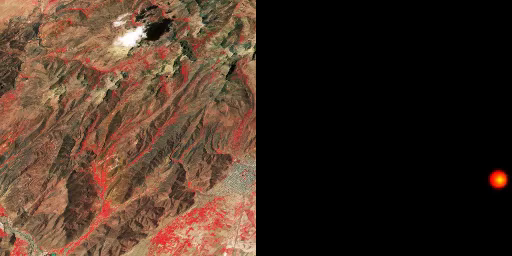}
    \includegraphics[width=0.23\linewidth]{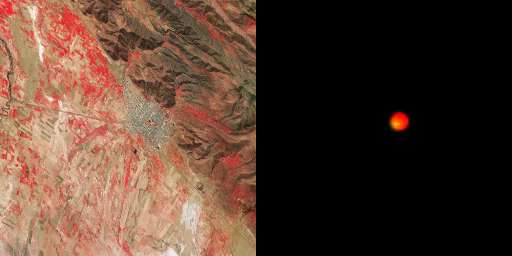}
    \includegraphics[width=0.23\linewidth]{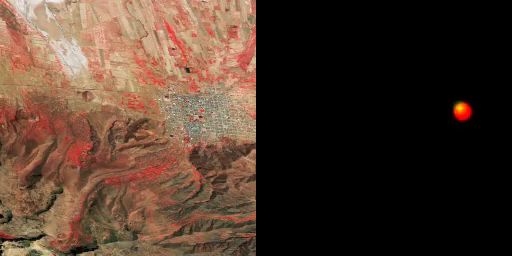}

    \includegraphics[width=0.23\linewidth]{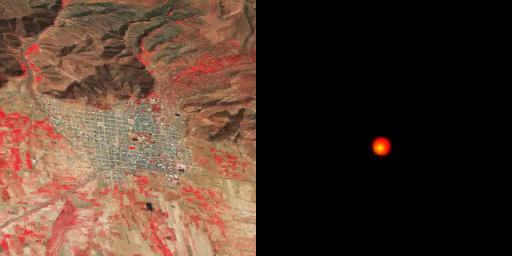}
    \includegraphics[width=0.23\linewidth]{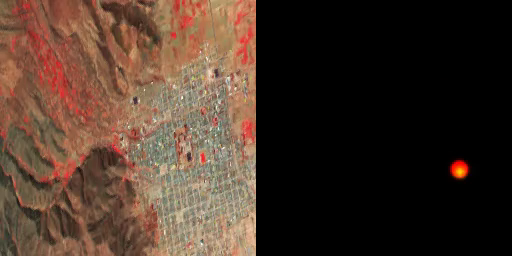}
    \includegraphics[width=0.23\linewidth]{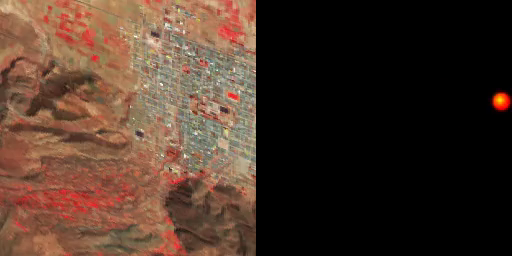}
    \includegraphics[width=0.23\linewidth]{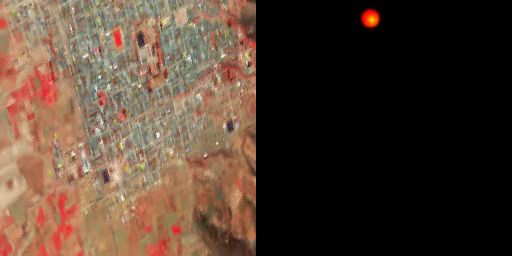}
    
    \caption{Outputs along a trajectory: all 8 images represent an image and an output mask (red dot is the location of the target, yellow is the pixel-wise predicted likelihood of being the target location). One can again notice the ground resolution difference between first en final image, yet the algorithm can coarsely predict the location of the target in all those situations.} 
    %\label{fig:enter-label}
\end{figure}

\section{Conclusion}
In this paper, we point out the limits of the registration-on-a-single-reference-image framework for sensor-based guidance and offer replacing it by directly learning a mapping between image and target localization using small deep convolutional networks on a stack of reference images.

\subsection*{Limits}
Despite successes in these preliminary experiments, it is obvious that this framework has many critical drawbacks compared to \textit{registration on one reference image}.
First, the offered framework expects a stack of reference images, increasing the burden of collecting data.
Then, the offered framework is to learn a complete but dedicated network for each given target during mission preparation (making mission preparation more fastidious).
Finally, given the purpose of this algorithm on a cyber-physical platform, simple statistical evaluation on a test set (and removal of well-understood geometric routines) may raise many questions.
Further research will be needed to strengthening these results and evaluating at larger scale the relevancy and safety of deep-learning-based guidance for such critical tasks and platforms.

\subsection*{Acknowledgment}
We thank ESA and IGN for publicly releasing S2 image and the BD Ortho.

NVIDIA Llama Nemotron has been used for improving English writing but no generative AI tools has been used for the first version of this paper.

\bibliographystyle{plain}
\bibliography{refs}
\end{document}